\documentclass[lettersize,journal]{IEEEtran}
\usepackage{amsmath,amsfonts}
\usepackage{algorithmic}
\usepackage{algorithm}
\usepackage{array}
\usepackage[caption=false,font=normalsize]{subfig}
\usepackage{subcaption}
\usepackage{textcomp}
\usepackage{stfloats}
\usepackage{url}
\usepackage{verbatim}
\usepackage{graphicx}
\usepackage{cite}
\usepackage{siunitx}
\usepackage{amsmath,amssymb,amsfonts}
\usepackage{algorithmic}
\usepackage{wrapfig}
\usepackage{booktabs}
\usepackage{threeparttable}

\begin{document}

\title{Geometry-Driven Opti-Acoustic Co-Registration and View-Invariant Reflectivity Mapping for Side-Scan Sonar}

\author{Taqi Hamoda,
        and Nuno Gracias,~\IEEEmembership{Member,~IEEE}

\thanks{All authors are with the Computer Vision and Robotics Research Institute (ViCOROB) of the University of Girona, Spain.}

\thanks{Corresponding author: Nuno Gracias (ngracias@silver.udg.edu ).}}



\maketitle

\begin{abstract}
Side-Scan Sonar (SSS) is a primary modality for large-scale underwater mapping, yet automated perception and cross-modal alignment are severely bottlenecked by acoustic complexities such as speckle noise, shadows, and extreme viewpoint dependencies. 
Traditional handcrafted descriptors and modern deep learning matchers fail to bridge the physical domain gap between optical and acoustic imagery without 3D geometric constraints. 
To overcome these limitations, we propose a novel geometry-driven framework for pixel-level opti-acoustic co-registration and view-invariant reflectivity mapping. 
Our method utilizes Structure-from-Motion (SfM) to reconstruct a dense 3D seafloor mesh, acting as a geometric anchor between the visual and acoustic domains. 
We introduce a First Bottom Return (FBR) extraction algorithm to dynamically correct non-linear altitude drift caused by uncalibrated SfM reconstruction. 
Furthermore, we apply an inverse Lambertian model and a dual-Gaussian weighting function to isolate the intrinsic seabed reflectivity, effectively neutralizing slant-range propagation loss and geometric view-dependence. 
By deterministically associating these isolated acoustic properties with optical pixels, our pipeline generates highly accurate, strictly co-registered multi-modal datasets. 
This automated, physics-guided approach eliminates the need for manual annotation and paves the way for advanced self-supervised learning in benthic habitat mapping.
\end{abstract}


\section{Introduction}

\IEEEPARstart{U}{nderwater} robotic perception is severely constrained by the physical properties of the marine environment. Due to rapid absorption and scattering of electromagnetic radiation, high-resolution optical sensors are typically limited to operational ranges of under 10 meters \cite{can2025_sss,katrusha2025}. 
Conversely, acoustic energy propagates with minimal attenuation, allowing sound waves to travel hundreds of meters even in turbid or light-deprived waters. Consequently, Side-Scan Sonar (SSS) has emerged as the primary sensing modality for large-scale seafloor mapping, target detection, and autonomous navigation \cite{antoni2016,hayat2025}.

\begin{figure}[t] 
\centering 
\includegraphics[width=\columnwidth]{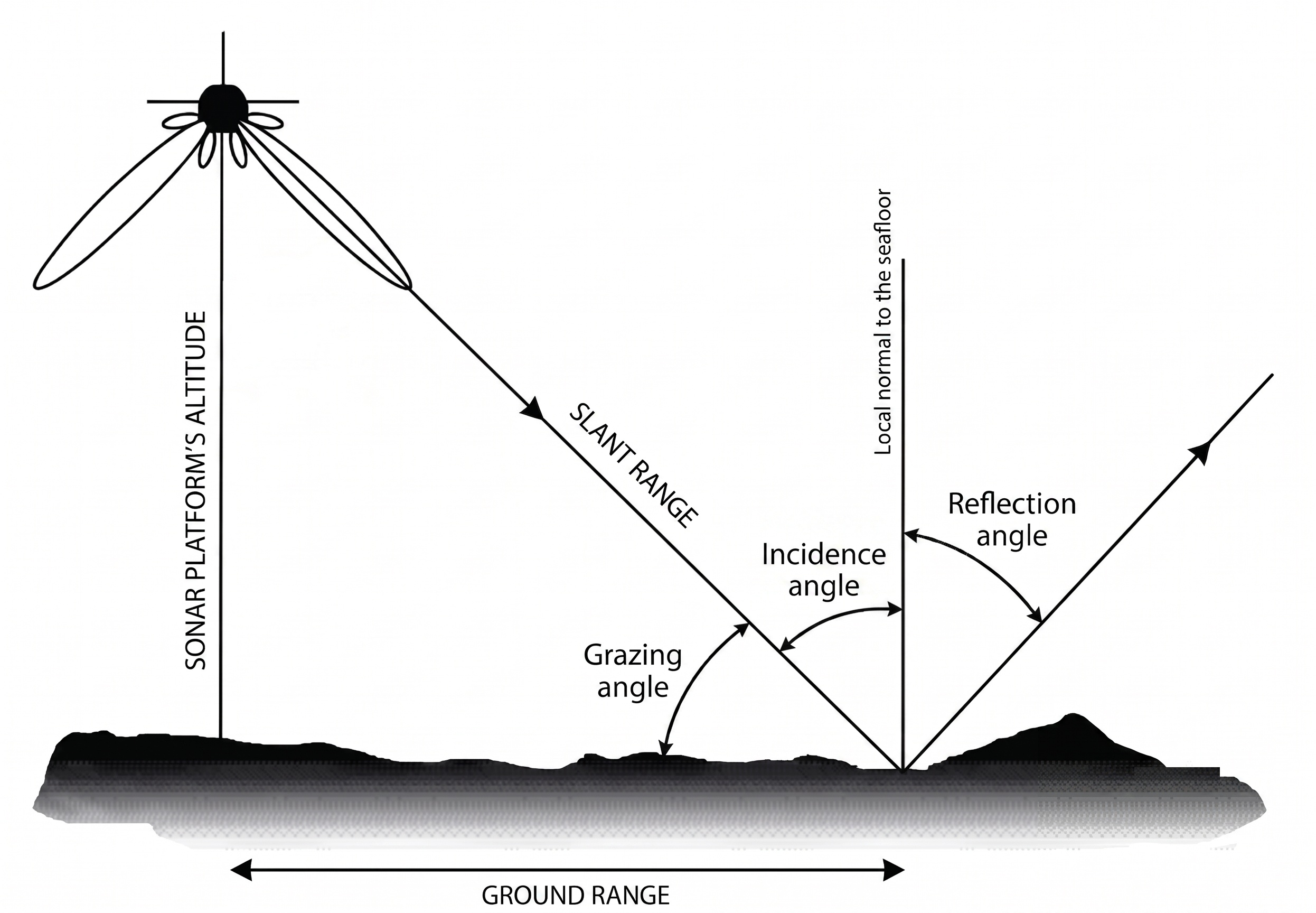} 
\caption{The Lambertian physics model for the SSS modality. Figure adapted from \cite{sss_handbook}} 
\label{fig:lambertian} 
\end{figure}

\begin{figure*}[t] 
\centering 
\includegraphics[width=\textwidth]{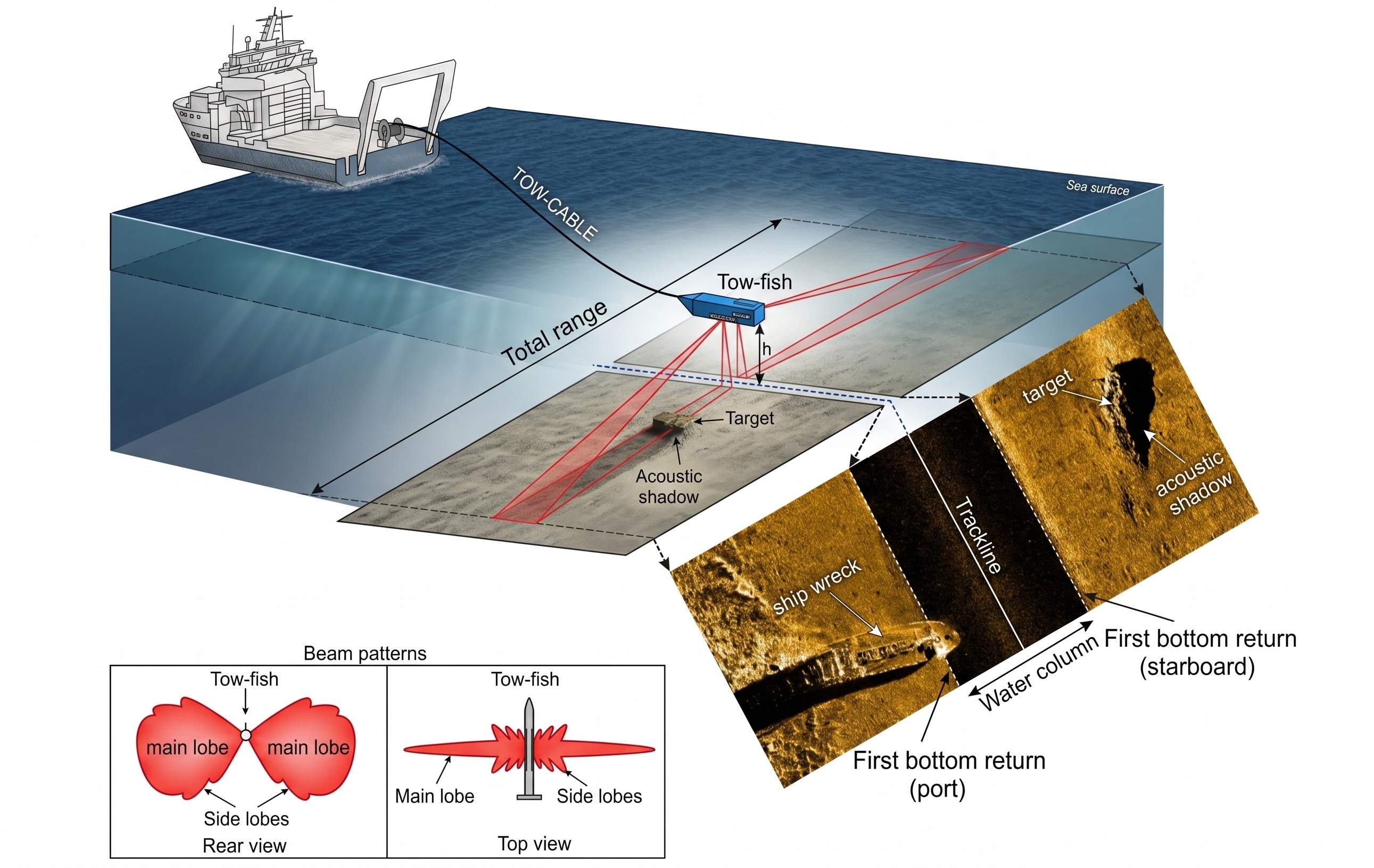} 
\caption{Side-Scan Sonar deployment on a towfish connected to a survey vessel. Figure adapted from \cite{sss_book}.} 
\label{fig:sss_towfish} 
\end{figure*}

Unlike optical cameras, SSS image formation is strictly governed by acoustic acquisition dynamics \cite{antoni2016, can2025_physdnet}. 
As illustrated in Figure \ref{fig:sss_towfish}, laterally mounted transducers on a moving platform emit fan-shaped acoustic pulses perpendicular to the vehicle's trajectory. The returning echoes are recorded over time and converted to range by assuming a constant sound speed of $1500\,\mathrm{m/s}$, yielding a continuous 2D backscatter intensity map \cite{antoni2016,hayat2025}. 
Traditionally, this process is approximated via a Lambertian reflection model: 
\begin{equation} 
I(x,y) = \rho(x,y) \cdot \cos(\theta(x,y)) \cdot L(x,y), 
\end{equation} 
where the recorded intensity $I(x,y)$ is a function of the intrinsic seabed reflectivity $\rho(x,y)$, the local incidence angle $\theta(x,y)$, and the acoustic propagation loss $L(x,y)$ \cite{can2025_physdnet,antoni2016}.

However, this simplified theoretical model fails to capture the true complexities of acoustic propagation. Real SSS transducers emit non-uniform power profiles that diminish at the beam edges and fluctuate across side lobes \cite{yvan_petillot}. 
Furthermore, topographic occlusions generate stark acoustic shadows, while multiplicative speckle noise inherently corrupts the returning signal \cite{katrusha2025,alan2022}. 
Consequently, SSS imagery is highly viewpoint-dependent; traversing the same seafloor structure from different survey directions yields drastically altered intensity patterns and shadow geometries.

\subsection{State of the Art in Feature Extraction and Matching}
These severe physical artifacts pose a significant bottleneck for automated perception. Traditional handcrafted keypoint extractors (e.g., SIFT, SURF, AKAZE) rely heavily on local intensity gradients, which are easily disrupted by acoustic speckle and view-dependent backscatter, leading to poor repeatability \cite{lowe2004,can2025_sss,fu2025}. 

To overcome this, recent approaches have transitioned to deep-learned detectors (e.g., SuperPoint) and Transformer-based matchers (e.g., LightGlue, LoFTR) \cite{katrusha2025,fu2025}. 
Despite their immense success in optical domains, recent acoustic benchmarks reveal severe performance degradation under wave distortion and stripe noise, exposing critical architectural trade-offs \cite{katrusha2025}:
\begin{itemize} 
\item \textbf{Sparse Methods (e.g., SuperPoint + LightGlue):} Maintain relative accuracy under noise but yield incredibly sparse correspondences, failing to provide the dense alignment required for precise mapping. 
\item \textbf{Dense Methods (e.g., LoFTR):} Achieve high spatial coverage but remain hyper-sensitive to low-contrast acoustic artifacts, resulting in prohibitive computational footprints and low inlier rates.
\end{itemize}

A foundational limitation is that these networks are typically deployed "out-of-the-box" using pre-trained optical weights \cite{katrusha2025}.  
Without domain-specific fine-tuning, they fail to generalize to acoustic physics, relying instead on 2D homographic approximations that disregard 3D seafloor topography. This domain gap is exacerbated in cross-modal alignment (e.g., pairing SSS with optical imagery). To enforce structural consistency across differing sensor physics, symmetric epipolar distance is often introduced:
\begin{equation} 
d(p_1,p_2) = \frac{(p_2^\top F p_1)^2}{|F p_1|^2} + \frac{(p_1^\top F^\top p_2)^2}{|F^\top p_2|^2}, 
\end{equation} 
where $F \in \mathbb{R}^{3 \times 3}$ is the fundamental matrix \cite{fu2025}. 
However, optimizing this constraint requires features that are invariant to the underlying physics of both modalities—a requirement that current optical models simply cannot fulfill.

\subsection{Contributions} 
To address acoustic degradation, viewpoint dependency, and the severe domain gap in underwater multi-modal perception, this paper introduces a physics-informed, geometry-driven framework for SSS imagery. The primary contributions of this work are:
\begin{itemize} 
\item \textbf{Geometric Opti-Acoustic Co-Registration:} We develop a novel cross-modal alignment framework that leverages reconstructed 3D optical geometry (SfM) as an anchor to achieve deterministic, pixel-level registration between visual and acoustic data.
\item \textbf{Dynamic Altitude Correction:} We introduce a First Bottom Return (FBR) extraction algorithm utilizing a specialized cost function to dynamically correct non-linear altitude drift and scale artifacts inherent in uncalibrated SfM reconstructions.
\item \textbf{Physics-Guided Reflectivity Isolation:} We formulate an inverse Lambertian model paired with a dual-Gaussian weighting function to successfully disentangle intrinsic seabed reflectivity from slant-range propagation loss and geometric view-dependence.
\item \textbf{Automated Dataset Generation:} We provide a scalable pipeline to generate perfectly co-registered opti-acoustic datasets, eliminating the need for manual acoustic annotation and enabling future self-supervised learning for benthic habitat mapping.
\end{itemize}

\section{Methodology}
To establish a robust pixel-level correspondence between side-scan sonar (SSS) and optical imagery, we propose a geometry-driven pipeline \cite{can2025_sss}. 
This framework reconstructs a shared 3D spatial representation to bridge the disparate sensor geometries, altitudes, and physical modalities. 

\subsection{Geometric Reconstruction and Calibration}
Following data acquisition via a dense, high-overlap lawnmower survey, we process the optical imagery using Vehicle Inertial Measurement Unit (IMU)-constrained Structure-from-Motion (SfM). 
This yields a precise 3D seafloor mesh that serves as the geometric ground truth.
However, the lack of a calibrated sensor model introduces non-linear altitude offsets relative to the seabed. 
These artifacts stem from unconstrained optimization behaviors within the SfM solver:
\begin{itemize}
    \item \textbf{Dynamic Focal Length Estimation:} Jointly optimizing the focal length induces an artificial zooming effect, causing spatial errors to propagate non-linearly away from the optical center.
    \item \textbf{Unconstrained Radial Distortion:} The solver minimizes reprojection errors locally but fails to model water-column-induced radial distortions, amplifying altitude offsets.
\end{itemize}

\subsection{Altitude Offset Correction via First Bottom Return}
To resolve these non-linear altitude offsets, we utilize First Bottom Return (FBR) extraction from the SSS waterfall imagery to determine the true Autonomous Underwater Vehicle (AUV) altitude \cite{antoni2016}. 
The raw sonar channels are smoothed using a median filter to reduce speckle noise, followed by Contrast Limited Adaptive Histogram Equalization (CLAHE) to enhance edge contrast. 
A greedy algorithm tracks the FBR by minimizing the following cost function over a 100-bin sliding window:
\begin{equation}
\mathcal{C}(b) = 0.4 \cdot \nabla(b) + 0.3 \cdot \mathcal{D}_{\text{nadir}}(b) + 0.3 \cdot \mathcal{D}_{\text{past}}(b)
\end{equation}
where $\nabla(b)$ is the gradient magnitude, $\mathcal{D}_{\text{nadir}}(b)$ is the spatial distance from the nadir, and $\mathcal{D}_{\text{past}}(b)$ is the distance relative to preceding FBR locations. 
The FBR-derived ground-truth altitude dynamically corrects the reconstructed 3D geometry using an adaptive temporal window of the 50 closest pings, successfully compensating for non-linear drift.

\subsection{Physics-Guided Reflectivity Isolation}
To extract the view-invariant material reflectivity, an inverse Lambertian reflection model is applied to the geometrically aligned SSS data and 3D mesh \cite{can2025_physdnet}. 
Mesh vertices with occluded lines-of-sight are classified as acoustic shadows via raytracing and discarded. 
Valid vertices are mapped to specific SSS bins, and raw intensities are divided by the cosine of the local incidence angle $\theta$ and normalized along-track to remove slant-dependent propagation loss.
Finally, relative reflectivity values are projected back onto the 3D mesh using a dual-Gaussian weighting function $\mathcal{W}(\theta, r_s)$:
\begin{equation}
\mathcal{W}(\theta, r_s) = \exp\left( - \frac{(\theta - \pi/4)^2}{2(\pi/16)^2} \right) \cdot \exp\left( - \frac{r_s^2}{2(12)^2} \right)
\end{equation}
This isolates the final view-invariant reflectivity $\rho_{v}$ as a weighted average, neutralizing the sonar's emission physics and angular sensitivity to yield highly accurate co-registered data.

\section{Results and Discussion}
Following the geometric and physics-based corrections detailed in our methodology, we evaluated the framework's performance across diverse benthic environments, specifically rocky and seagrass regions.

\subsection{Altitude Correction Performance}
\begin{figure}[t] 
    \centering 
    \subfloat[]{%
        \includegraphics[width=0.48\columnwidth]{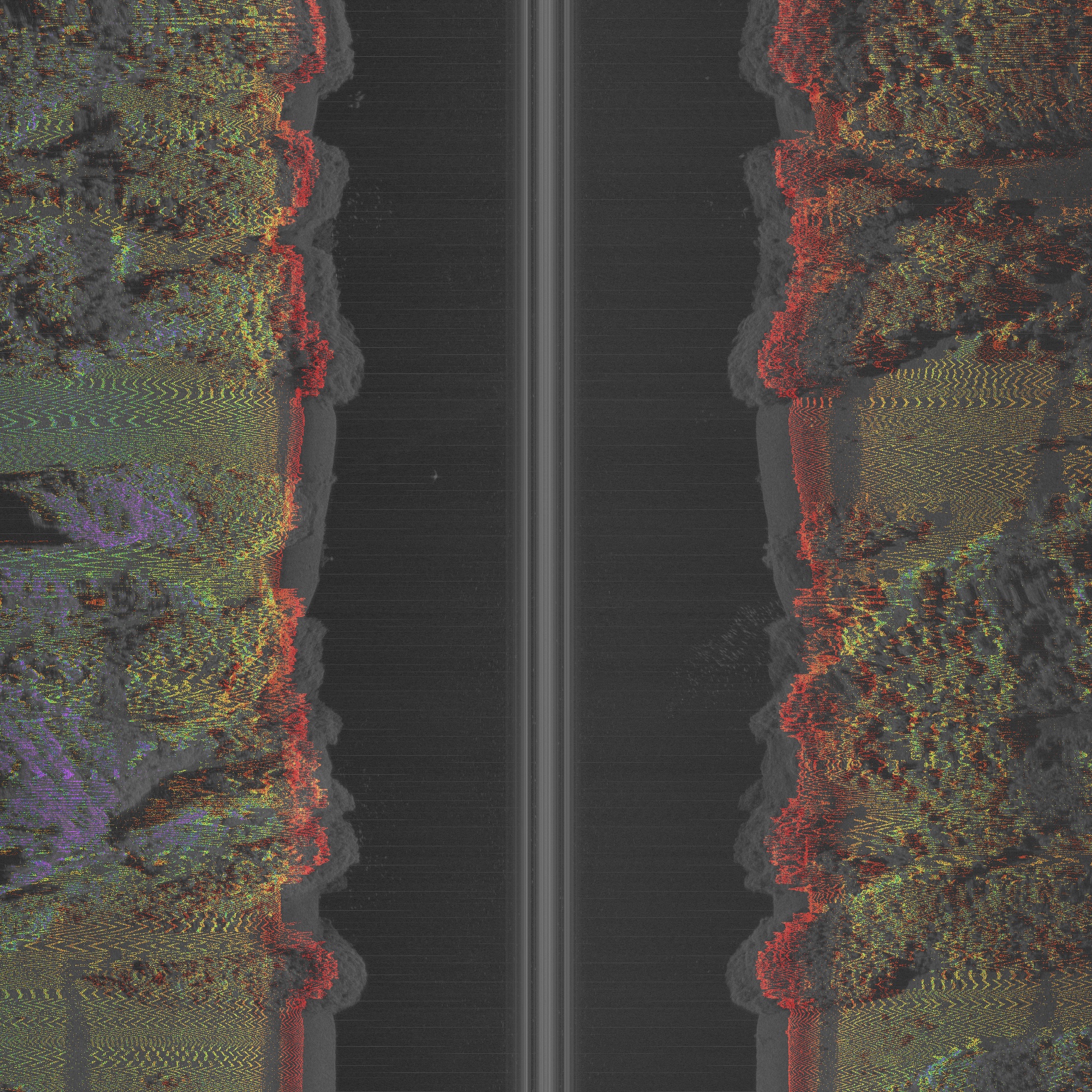}%
        \label{fig:offset_before}%
    } \hfill 
    \subfloat[]{%
        \includegraphics[width=0.48\columnwidth]{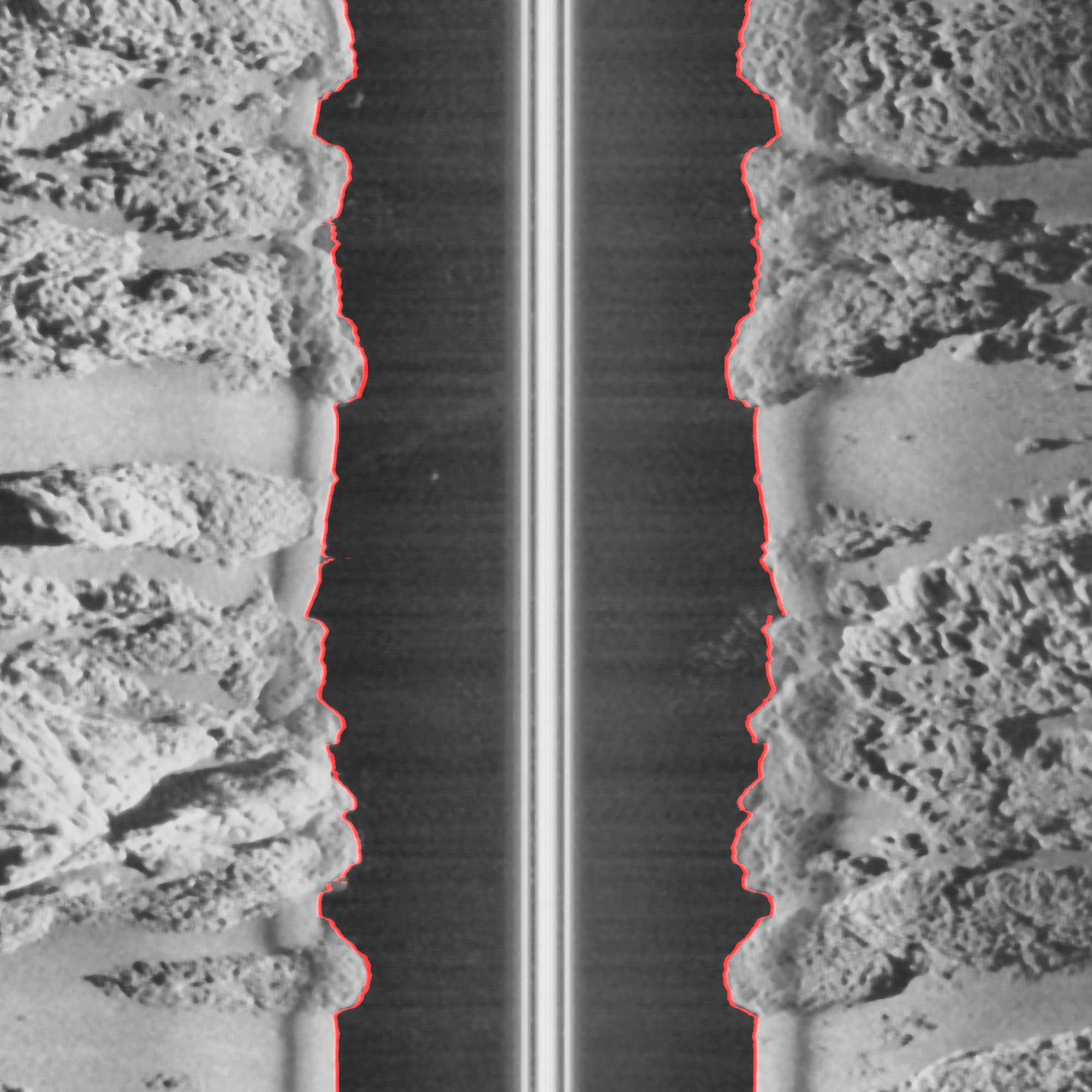}%
        \label{fig:offset_fbr}%
    } \\ 
    \subfloat[]{%
        \includegraphics[width=0.48\columnwidth]{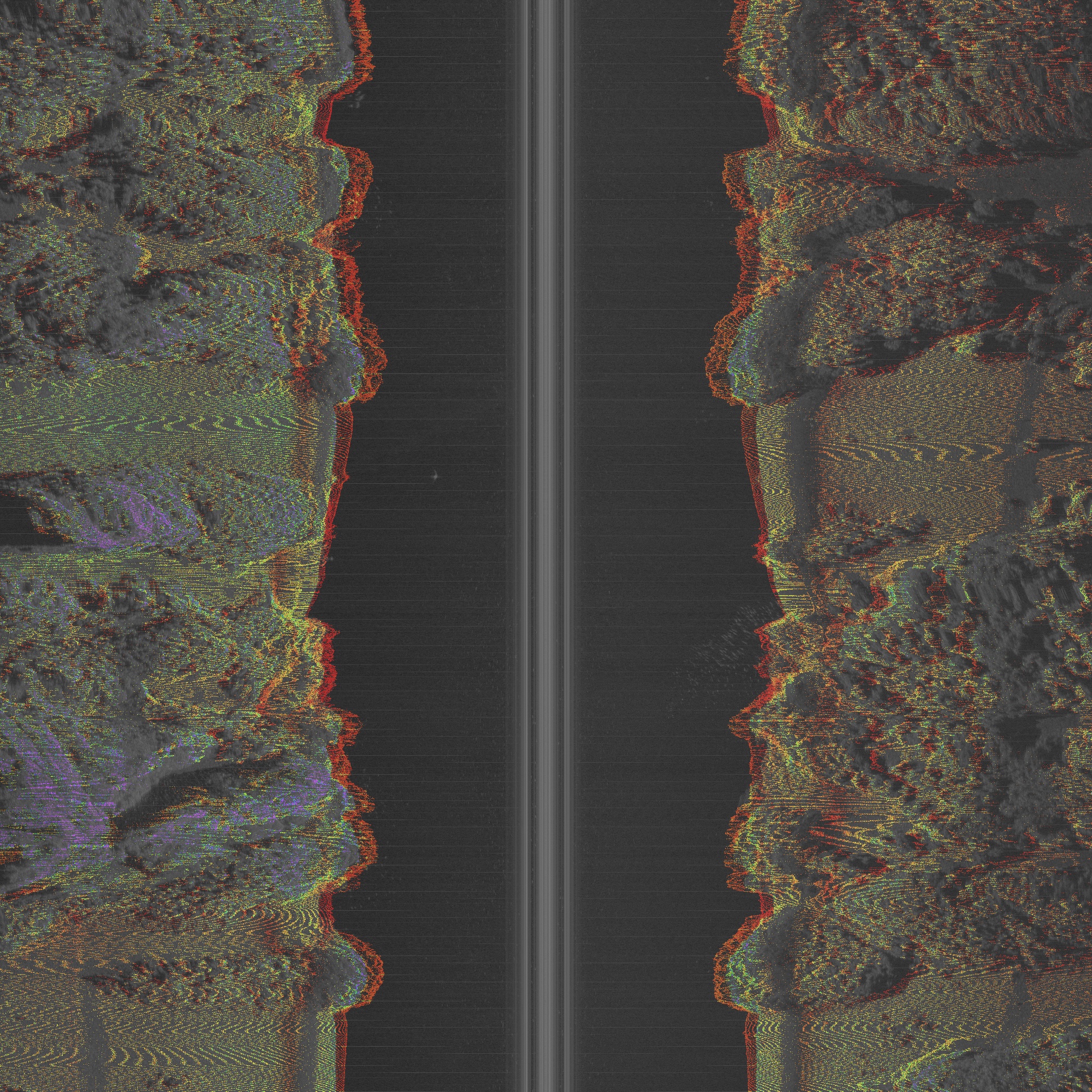}%
        \label{fig:offset_global}%
    } \hfill 
    \subfloat[]{%
        \includegraphics[width=0.48\columnwidth]{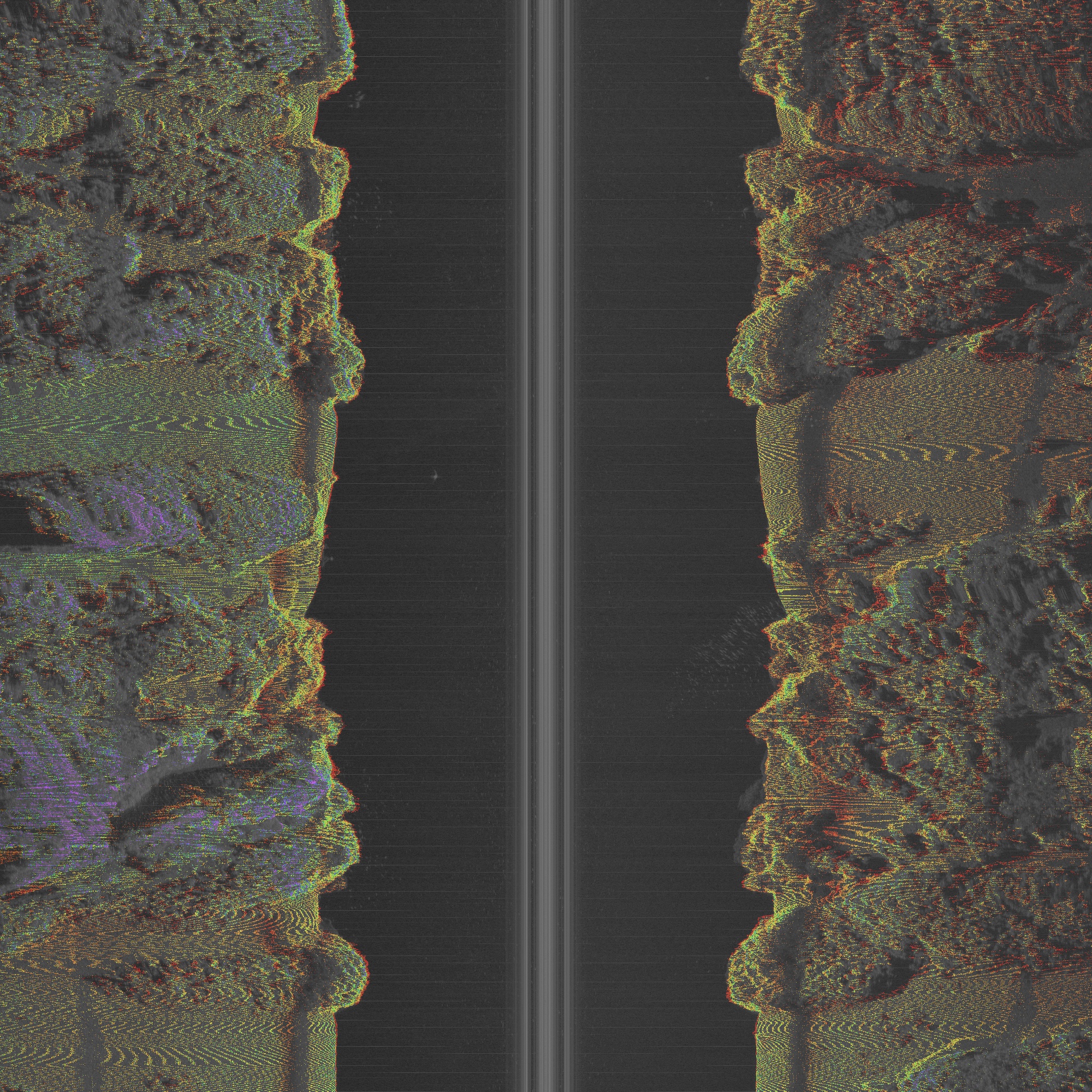}%
        \label{fig:offset_local}%
    } 
    \caption{Results of the offset correction pipeline: (a) offset between SSS waterfall and geometry reprojection, (b) First Bottom Return (FBR) detection performance, (c) after global offset correction, and (d) after local offset correction. OpenCV's Rainbow colormap is applied to the relative acoustic reflectivity.} 
    \label{fig:results_offset} 
\end{figure}

Uncalibrated Structure-from-Motion (SfM) reconstructions inherently introduce non-linear altitude drift between the estimated vehicle pose and the seabed \cite{can2025_sss}. 
However, the application of our dynamic, local First Bottom Return (FBR) offset correction successfully compensated for these non-linear scaling artifacts. 
As demonstrated in Figure \ref{fig:results_offset}, the local temporal correction achieved precise structural alignment between the raw SSS waterfall imagery and the corresponding geometric reprojection, vastly outperforming global offset adjustments.

\begin{figure}[t] 
    \centering 
    \subfloat[]{%
        \includegraphics[width=0.48\columnwidth]{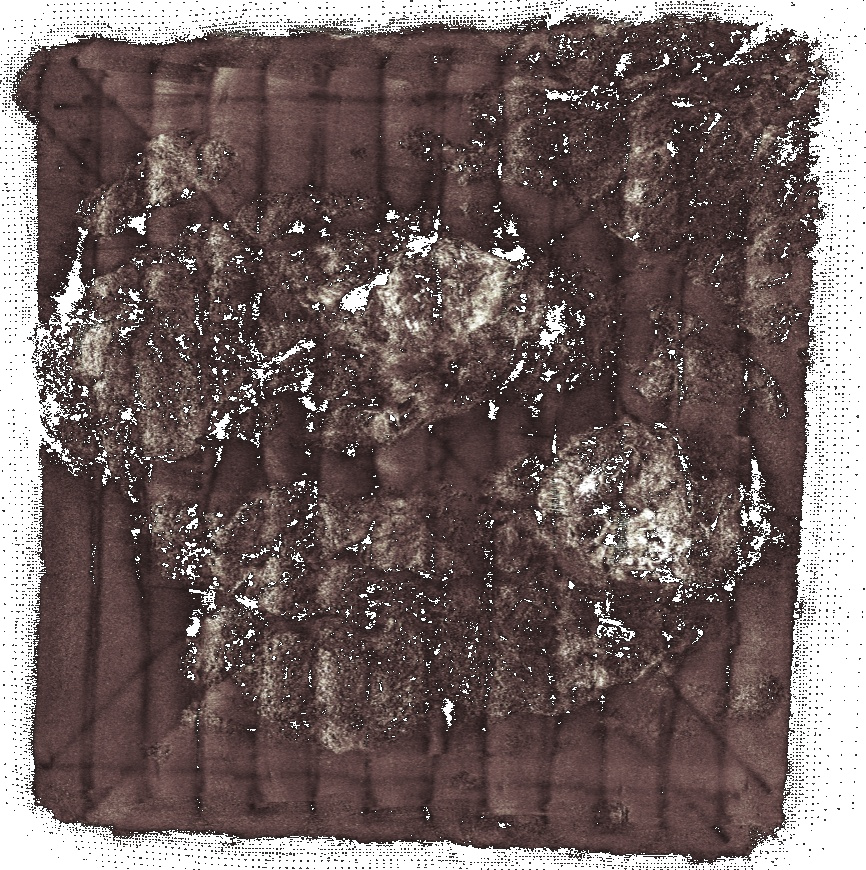}%
        \label{fig:ref_rocks}%
    } \hfill 
    \subfloat[]{%
        \includegraphics[width=0.48\columnwidth]{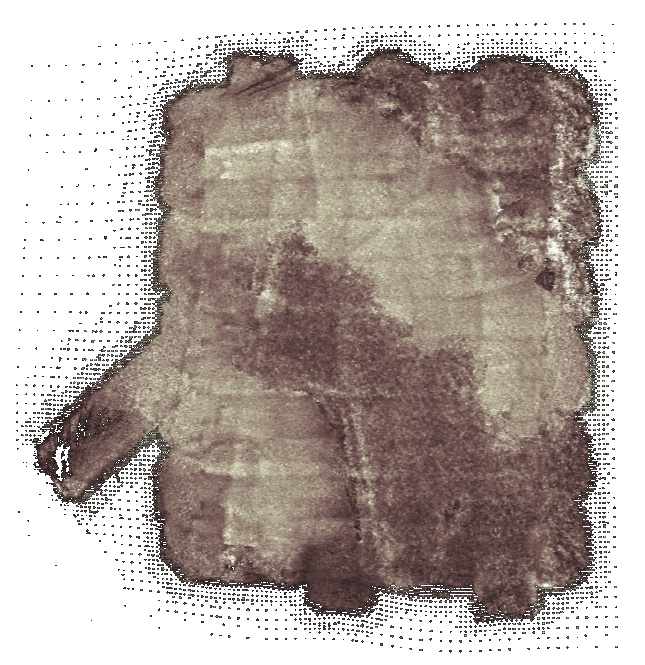}%
        \label{fig:ref_seagrass}%
    } 
    \caption{Estimated seabed reflectivity maps: (a) mountainous region mesh and (b) seagrass region mesh. Brighter areas indicate higher acoustic reflectivity.} 
    \label{fig:results_reflectivity} 
\end{figure}

\subsection{Reflectivity Mapping Outcomes}
\begin{figure}[t] 
    \centering 
    \subfloat{%
        \includegraphics[width=0.31\columnwidth]{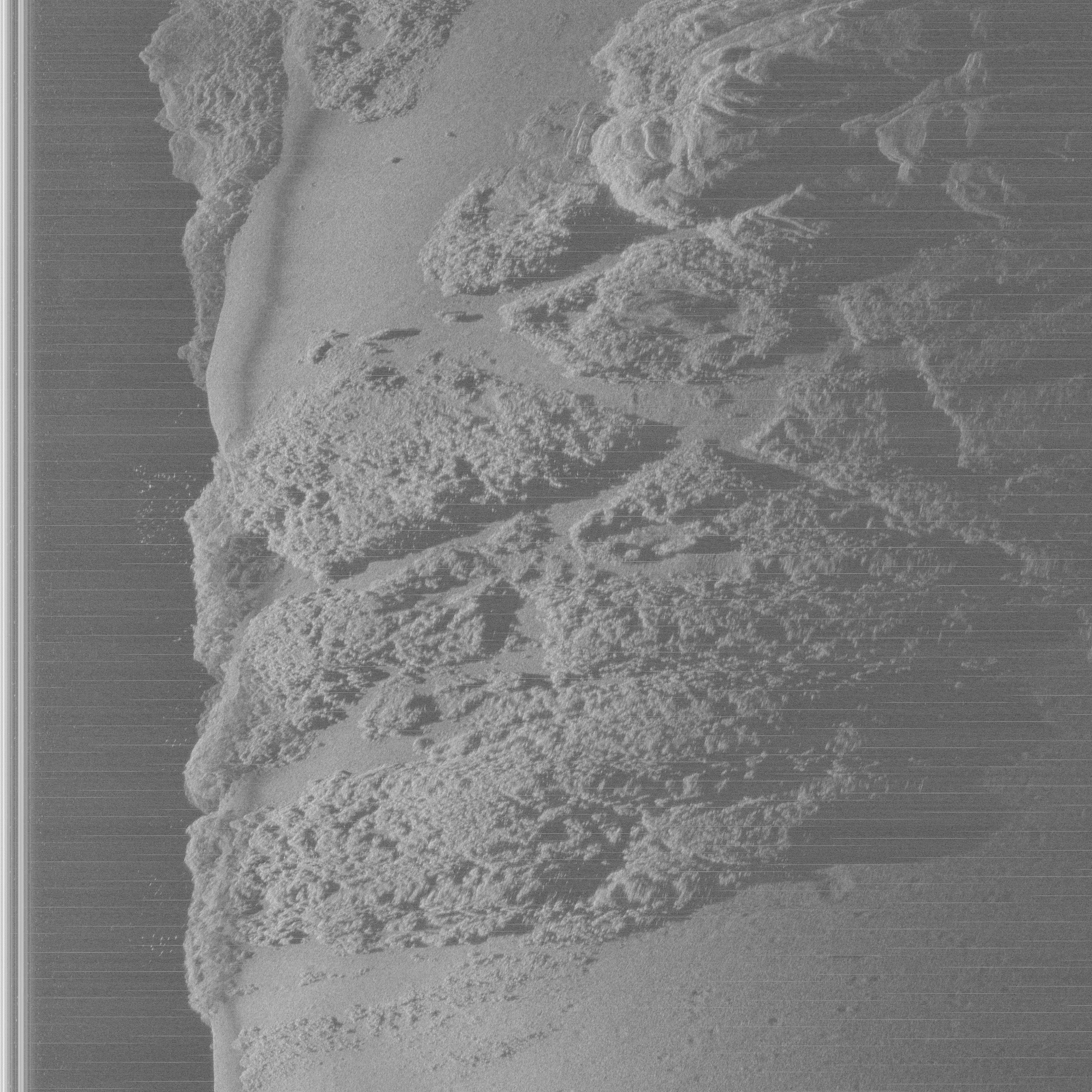}%
        \label{fig:synth_orig}%
    } \hfill 
    \subfloat{%
        \includegraphics[width=0.31\columnwidth]{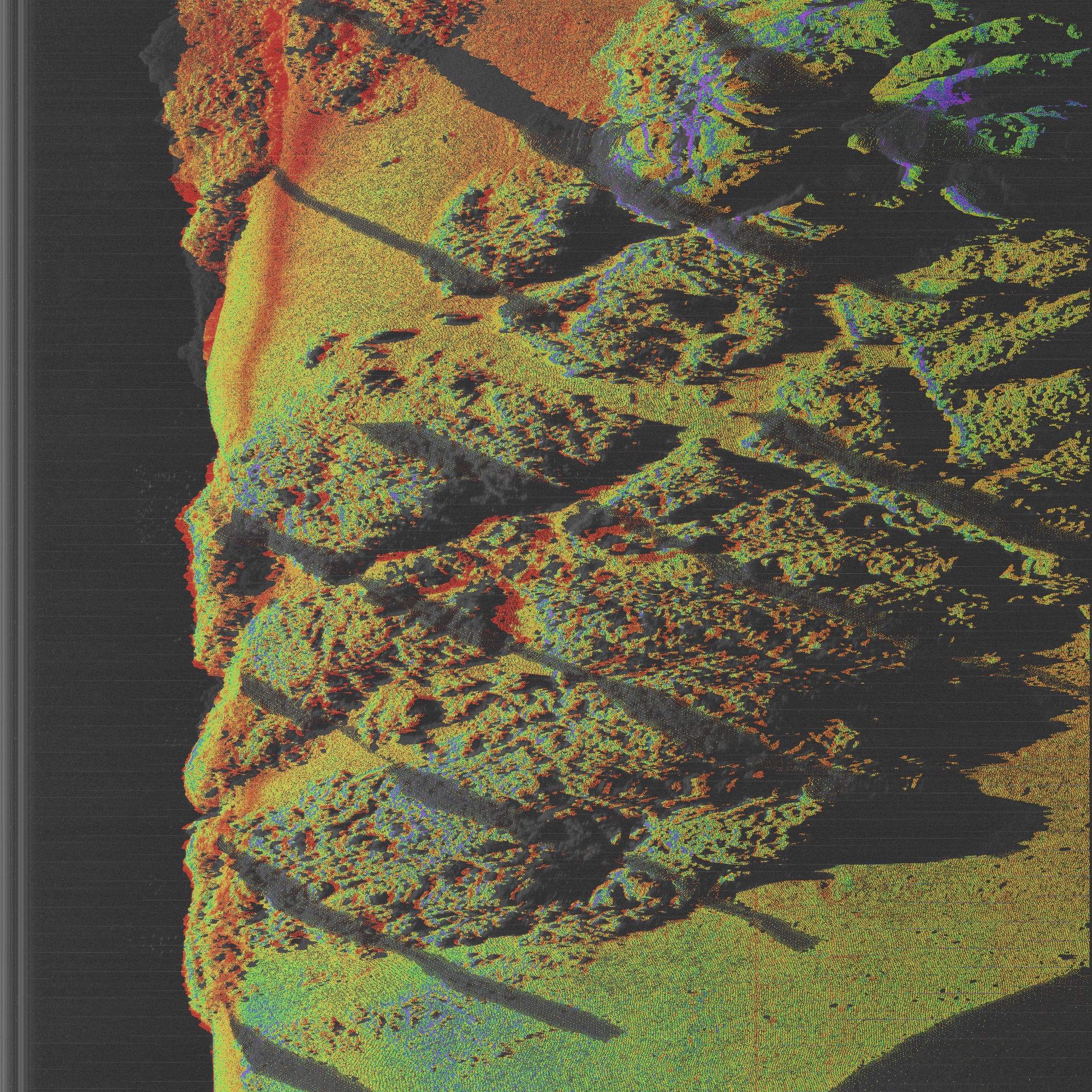}%
        \label{fig:synth_over}%
    } \hfill 
    \subfloat{%
        \includegraphics[width=0.31\columnwidth]{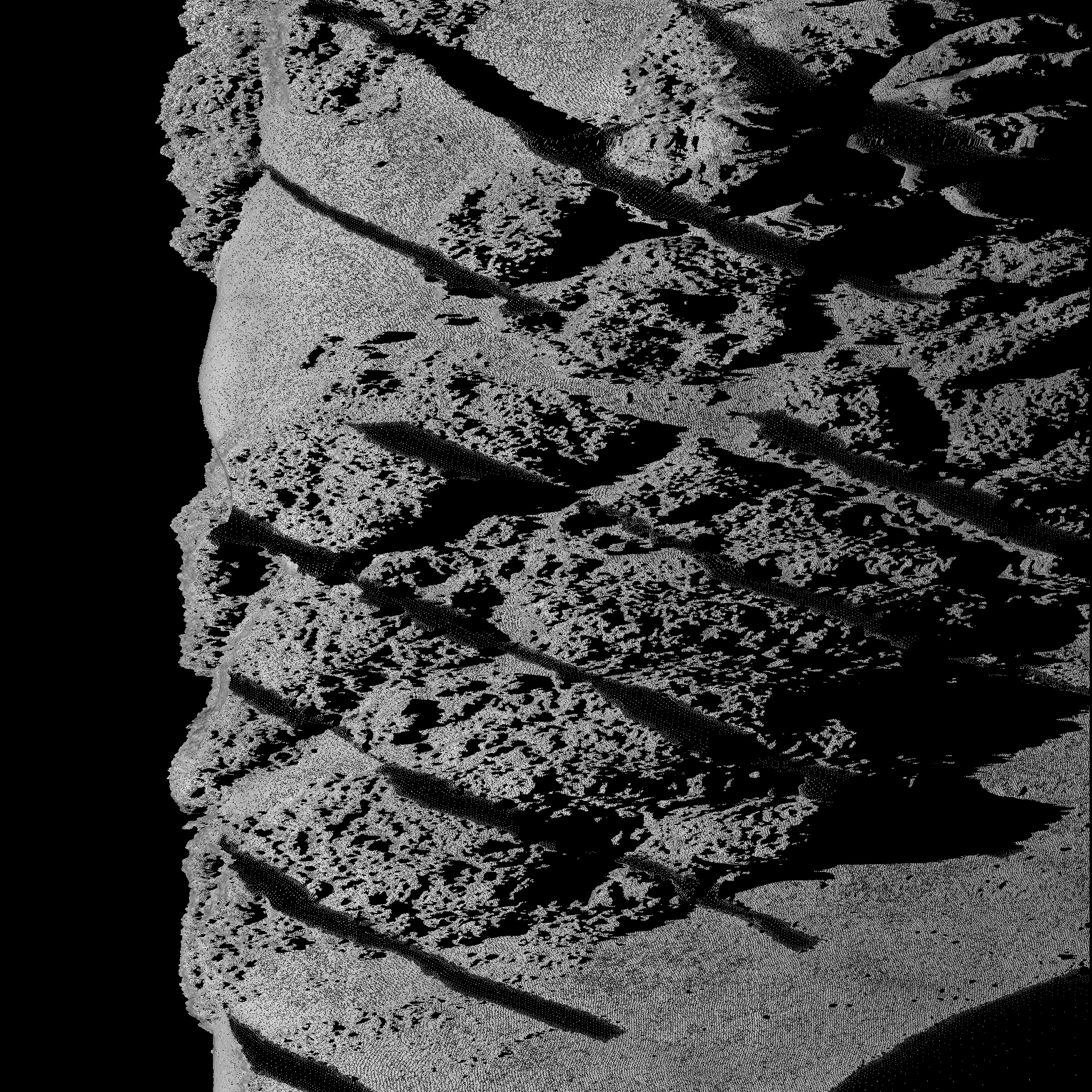}%
        \label{fig:synth_res}%
    } \\
    \subfloat{%
        \includegraphics[width=0.31\columnwidth]{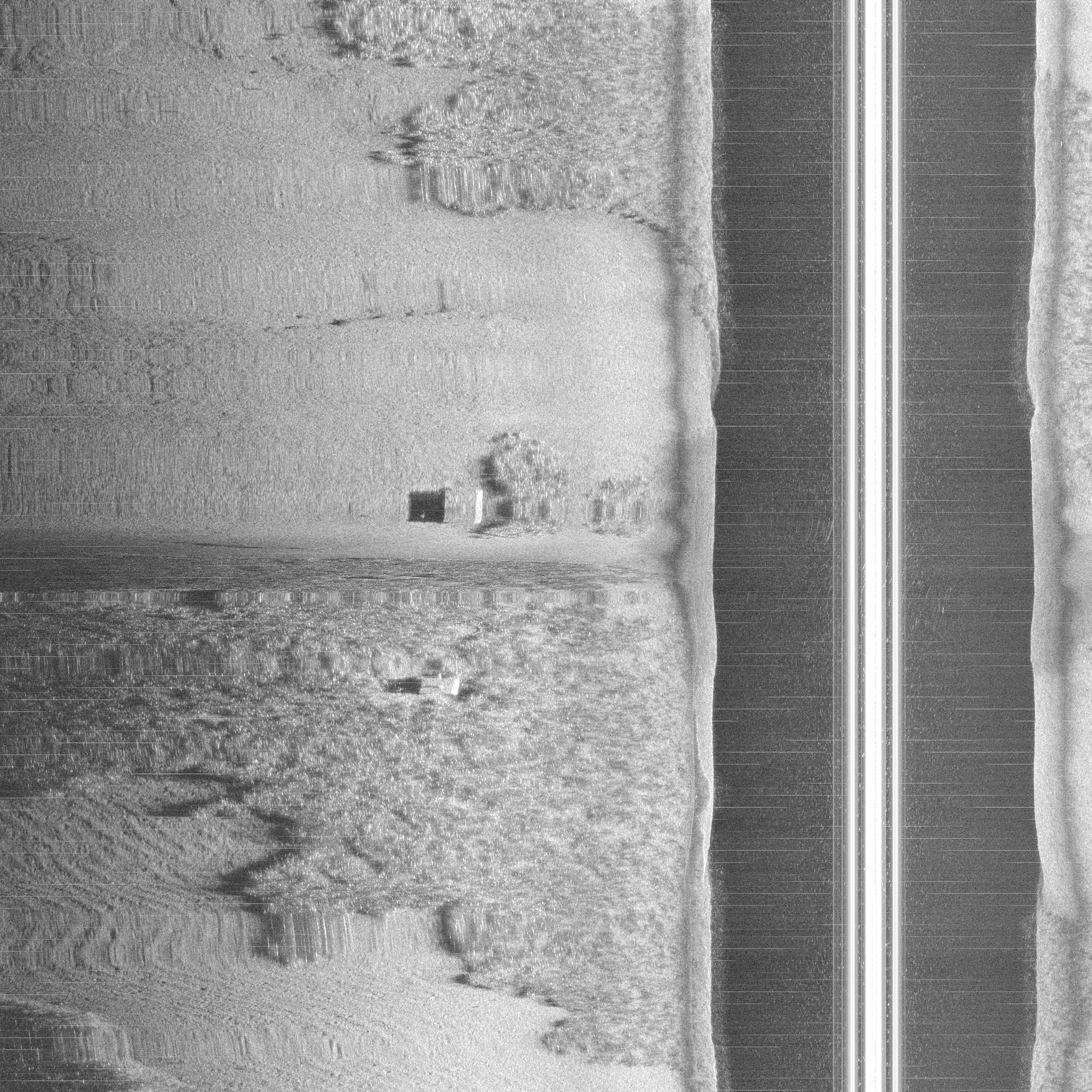}%
        \label{fig:synth_orig}%
    } \hfill 
    \subfloat{%
        \includegraphics[width=0.31\columnwidth]{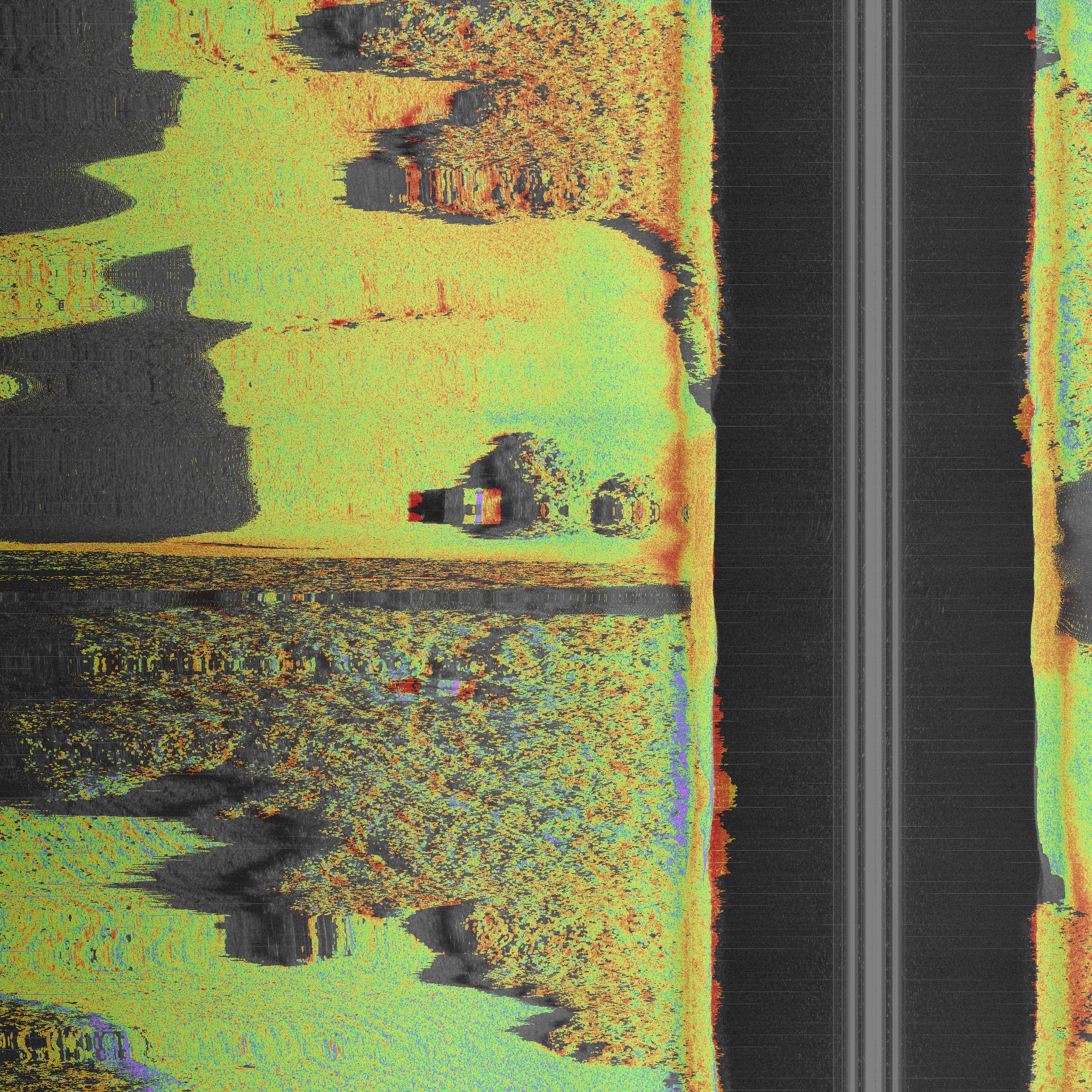}%
        \label{fig:synth_over}%
    } \hfill 
    \subfloat{%
        \includegraphics[width=0.31\columnwidth]{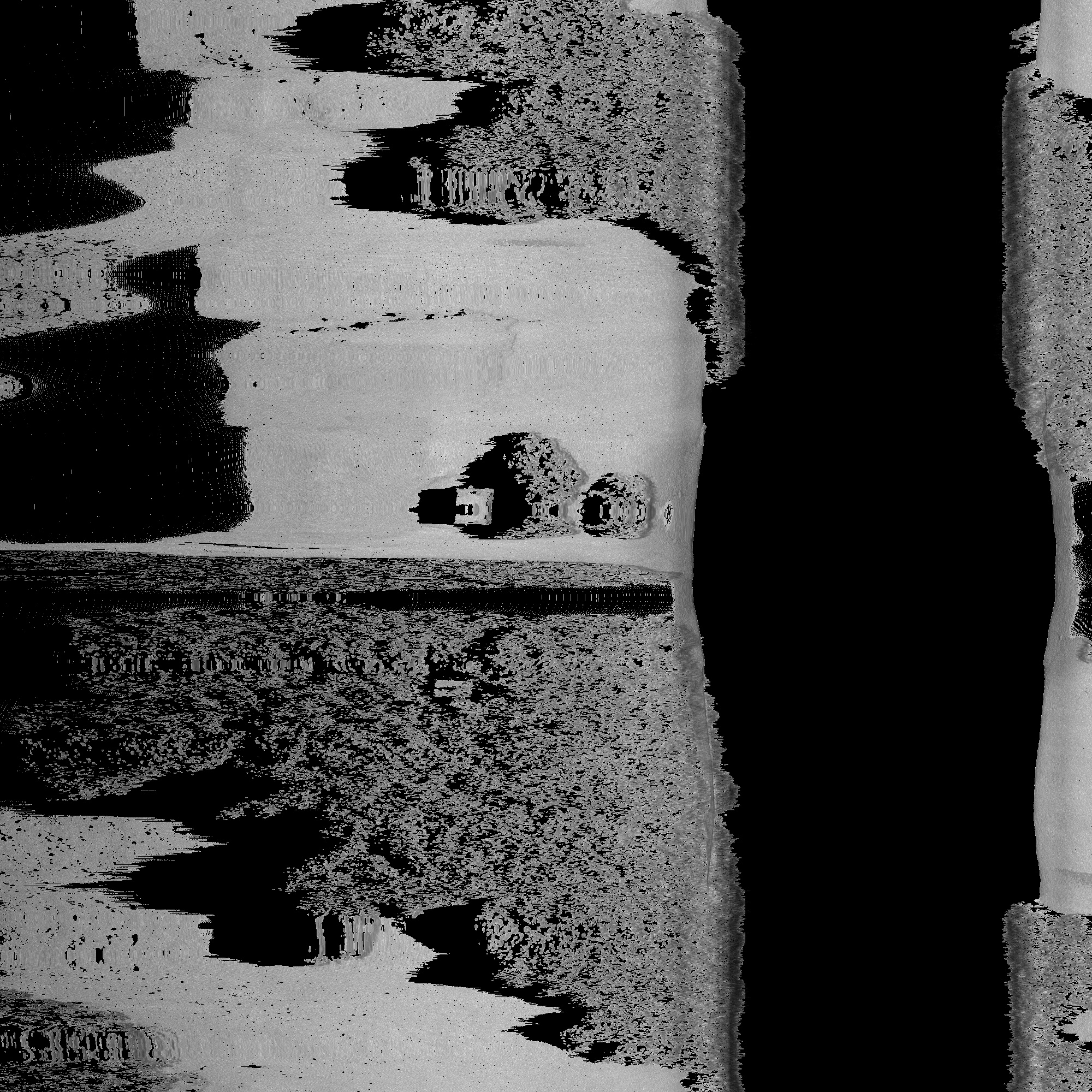}%
        \label{fig:synth_res}%
    }
    \caption{Acoustic simulation pipeline: (left) patch from the SSS waterfall, (middle) corresponding geometric reprojection, and (right) synthesized intensities based on the isolated reflectivity.} 
    \label{fig:results_synthetic} 
\end{figure}

By applying the inverse Lambertian reflection model, the pipeline successfully isolated the view-invariant material reflectivity ($\rho_v$) from transient acoustic artifacts \cite{can2025_physdnet}. 
Figure \ref{fig:results_reflectivity} illustrates the estimated seabed reflectivity maps, demonstrating the removal of geometric view-dependence and slant-dependent propagation loss. 
The integration of shadow filtering and dual-Gaussian blending ensured that the final mapped intensities accurately represent intrinsic seabed properties. 
Furthermore, these isolated reflectivity values facilitated the accurate synthesis of SSS intensities for novel-view acoustic simulation (Figure \ref{fig:results_synthetic}).

\subsection{Opti-Acoustic Co-Registration Results}
\begin{figure*}[t] 
    \centering 
    \subfloat[]{%
        \includegraphics[width=0.32\textwidth]{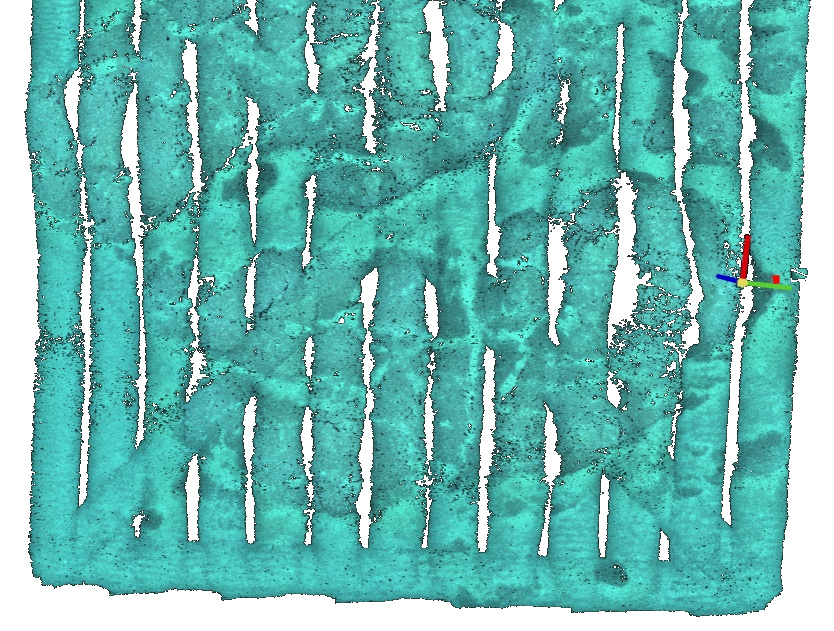}%
        \label{fig:reg_rocks_pcd}%
    } \hfill 
    \subfloat[]{%
        \includegraphics[width=0.32\textwidth]{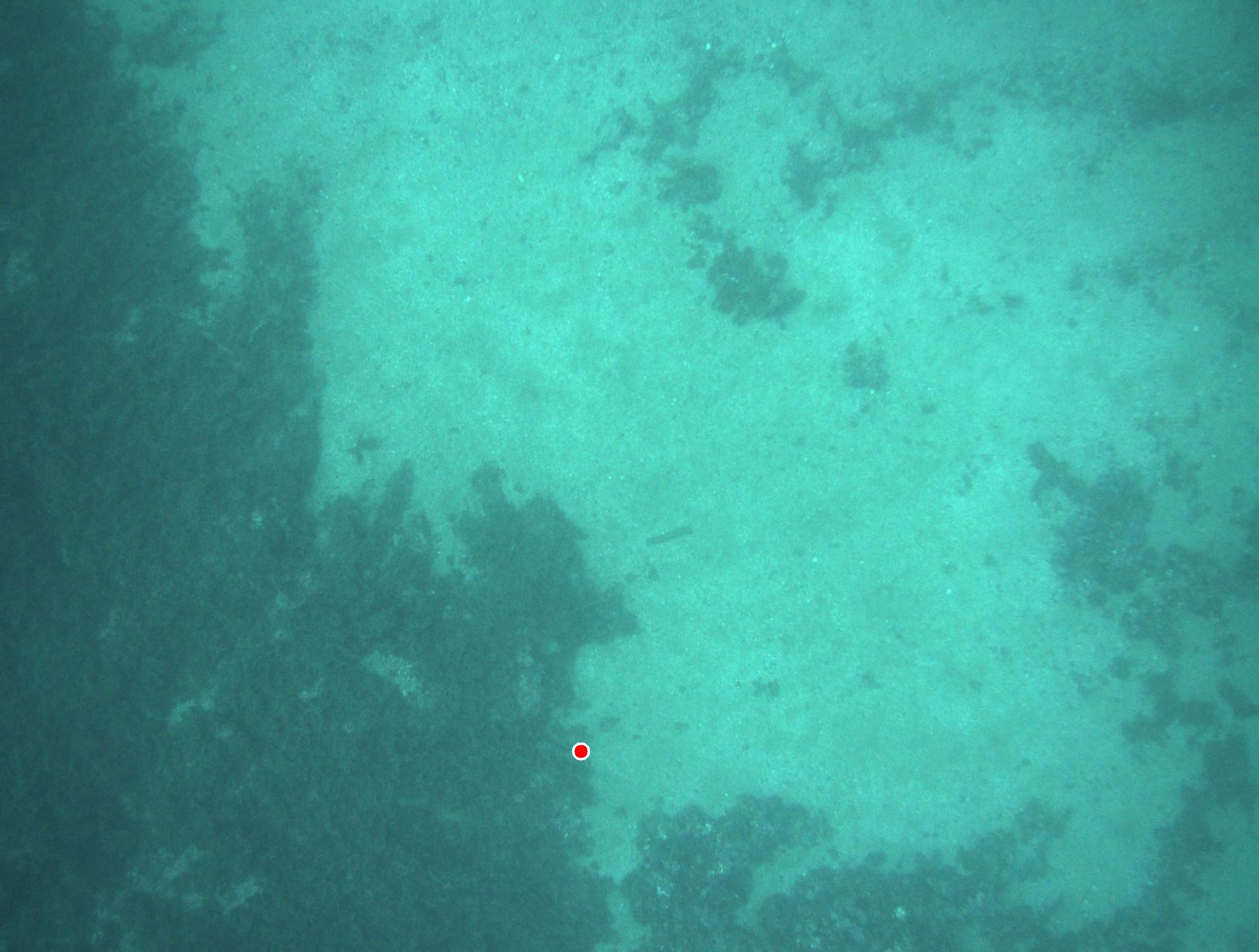}%
        \label{fig:reg_rocks_opt}%
    } \hfill 
    \subfloat[]{%
        \includegraphics[width=0.32\textwidth]{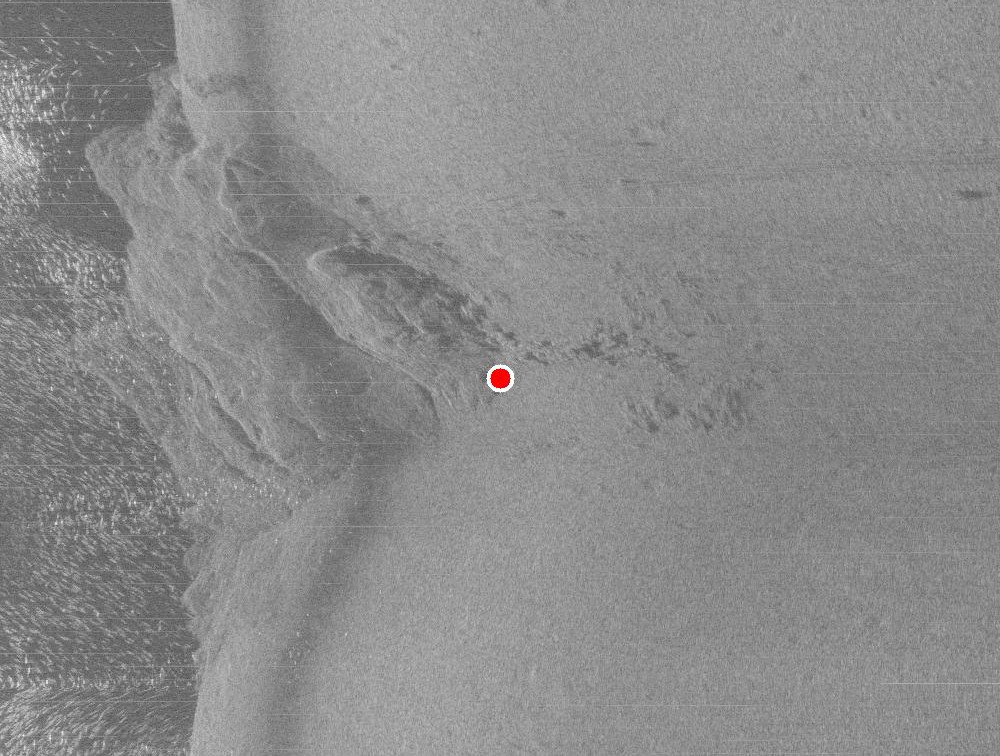}%
        \label{fig:reg_rocks_sss}%
    } \\[4pt] 
    \subfloat[]{%
        \includegraphics[width=0.32\textwidth]{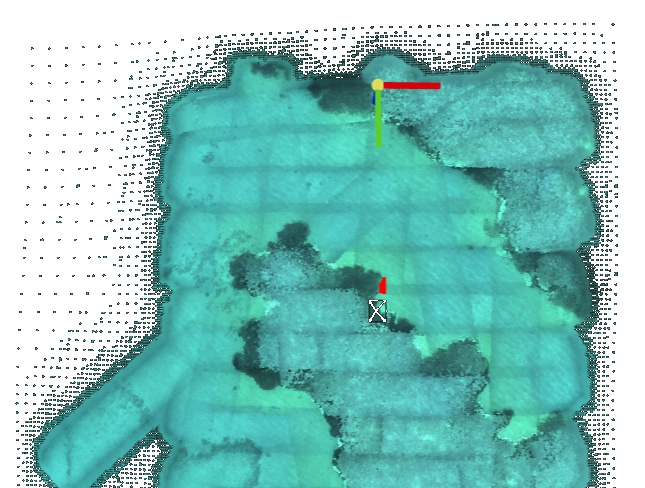}%
        \label{fig:reg_seagrass_pcd}%
    } \hfill 
    \subfloat[]{%
        \includegraphics[width=0.32\textwidth]{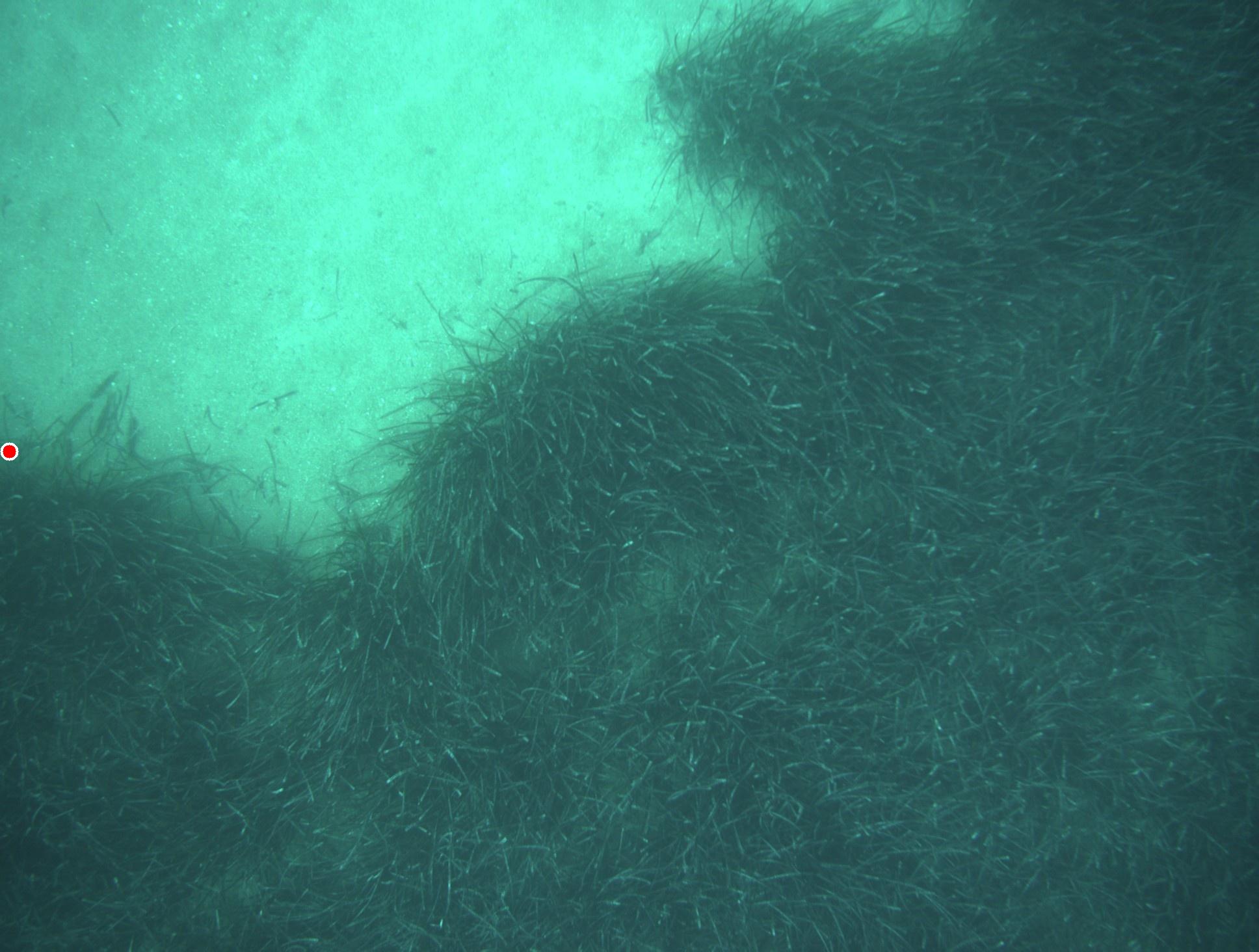}%
        \label{fig:reg_seagrass_opt}%
    } \hfill 
    \subfloat[]{%
        \includegraphics[width=0.32\textwidth]{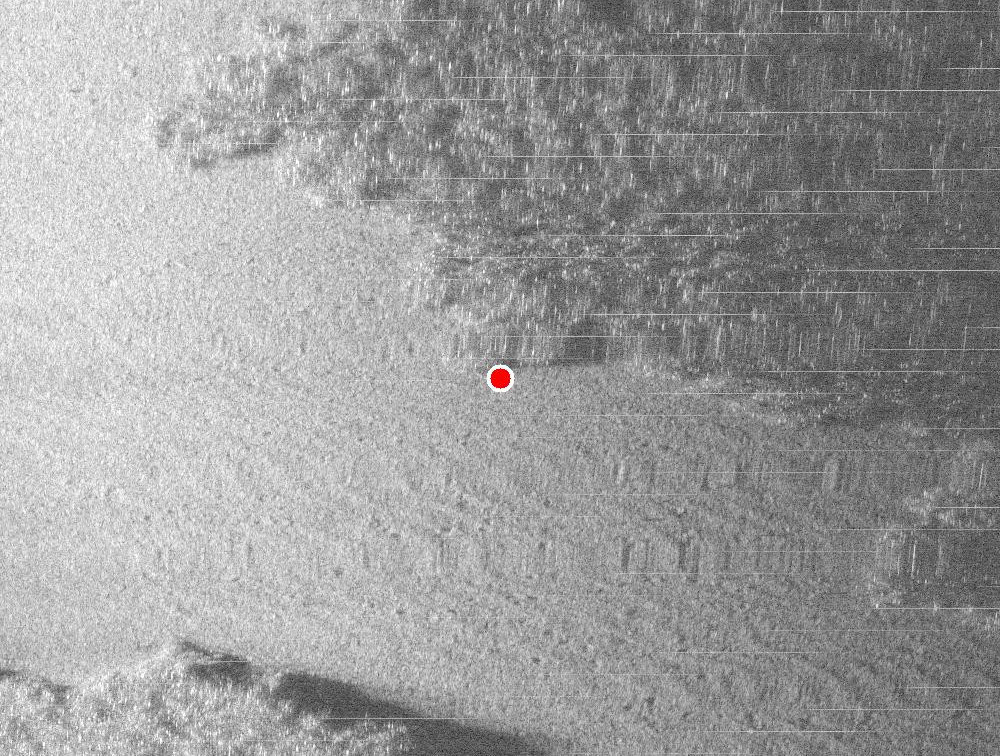}%
        \label{fig:reg_seagrass_sss}%
    }
    \caption{Geometric opti-acoustic co-registration results across two test regions. Top row (rocky seabed): (a) 3D point cloud, (b) optical image, and (c) SSS waterfall. Bottom row (seagrass bed): (d) 3D point cloud, (e) optical image, and (f) SSS waterfall.} 
    \label{fig:results_registration} 
\end{figure*}

The spatial bridge established between localized acoustic bins and the 3D SfM point cloud achieved a highly accurate, deterministic pixel-level co-registration. 
Figure \ref{fig:results_registration} showcases this successful cross-modal alignment across both the mountainous and seagrass test regions.  
By eliminating spatial ambiguity in multi-modal sensor fusion, this automated pipeline yields a high-fidelity proxy for ground truth. 
Ultimately, this generated dataset provides the foundational data necessary to train Self-Supervised Learning (SSL) architectures for benthic classification, entirely bypassing the need for manual acoustic annotation.

\section{Conclusion and Future Work}
This paper presented a novel geometry-driven and physics-informed framework to overcome the severe domain gap in opti-acoustic co-registration. While traditional handcrafted extractors and modern deep-learning matchers struggle with acoustic scattering, speckle noise, and extreme geometric distortions \cite{lowe2004, katrusha2025}, 
our approach successfully mitigates these challenges. By reconstructing a dense 3D SfM mesh as a structural anchor and dynamically correcting non-linear altitude drift via First Bottom Return (FBR) extraction, we established a deterministic pixel-level mapping between the visual and acoustic domains.
Furthermore, the integration of an inverse Lambertian physics model successfully disentangled viewpoint-dependent artifacts from intrinsic seabed reflectivity, yielding highly accurate, view-invariant co-registration \cite{can2025_physdnet, yvan_petillot}.

\subsection{Future Work} 
Future work will focus on rigorous quantitative evaluation and framework refinement through three primary avenues:
\begin{itemize} 
\item \textbf{Advanced Acoustic Modeling:} Framework fidelity can be improved by replacing the simplified Lambertian approximation with advanced acoustic scattering models and bathymetric priors to better capture complex, heterogeneous seabeds \cite{can2025_physdnet, yvan_petillot}. 
\item \textbf{Synthetic Benchmarking:} Utilizing synthetic data generated from the isolated reflectivity values will provide a rigorous benchmark to quantitatively evaluate the framework's novel-view acoustic simulation capabilities.
\item \textbf{Self-Supervised Learning Integration:} We plan to leverage the automatically generated, perfectly co-registered datasets to train Self-Supervised Learning (SSL) architectures for automated benthic habitat mapping, fully bypassing the bottleneck of manual acoustic annotation.
\end{itemize}

\section*{Acknowledgments}

This work was supported by Spanish Government through the project "Automated Seabed Analysis through Self-Supervised Deep Learning Sonar Technology (ASSiST)" under grant PID2023-149413OB-I00. 

\bibliographystyle{IEEEtran}
\bibliography{bibliography}

\end{document}